\documentclass[letterpaper, 10 pt, conference]{ieeeconf}  

\IEEEoverridecommandlockouts                              

\usepackage{graphicx} 
\usepackage{amsmath,amssymb,amsfonts}
\usepackage{hyperref}
\usepackage{siunitx}
\usepackage{float}
\usepackage{subcaption}

\usepackage{color}

\usepackage{array}
\newcolumntype{C}[1]{>{\centering\let\newline\\\arraybackslash}m{#1}}

\usepackage{booktabs}
\title{\LARGE \bf
FootstepNet: an Efficient Actor-Critic Method for Fast On-line Bipedal Footstep Planning and Forecasting
}

\author{Clément Gaspard$^{1*}$, Grégoire Passault$^{1*}$, Mélodie Daniel$^{1}$, Olivier Ly$^{1}$
\thanks{$^{1}$Univ. Bordeaux, CNRS, LaBRI, UMR 5800, 33400 Talence, France. Corresponding author: Clément Gaspard, e-mail: \texttt{clement.gaspard@u-bordeaux.fr}.
\newline This study has received financial support from the French government in the framework of the France 2030 program, Initiative of Excellence (IdEx) University of Bordeaux / RRI ROBSYS and from the ENS Rennes.
\newline $^{*}$ These authors equally contributed.}}

\begin{document}

\maketitle
\thispagestyle{empty}
\pagestyle{empty}

\begin{abstract}

Designing a humanoid locomotion controller is challenging and classically split up
in sub-problems. Footstep planning is one of those, where the sequence
of footsteps is defined. Even in simpler environments, finding a
\textit{minimal} sequence, or even a \textit{feasible} sequence, yields a complex optimization problem.
%
In the literature, this problem is usually addressed by search-based algorithms (\textit{e.g.} variants of A*). However, such approaches are either computationally expensive or rely on hand-crafted tuning of several parameters.
In this work, at first, we propose an efficient \textit{footstep planning} method to navigate in local environments with obstacles, based on state-of-the art Deep Reinforcement Learning (DRL) techniques, with very low
computational requirements for on-line inference. Our approach is heuristic-free and relies on a continuous set of actions to generate feasible footsteps. In contrast, other methods necessitate the selection of a relevant discrete set of actions.
%
Second, we propose a \textit{forecasting} method, allowing to quickly estimate the number of footsteps required
to reach different candidates of local targets. This approach relies on inherent computations made by the
\textit{actor-critic} DRL architecture.
We demonstrate the validity of our approach with simulation results, and by a deployment on a kid-size
humanoid robot during the RoboCup 2023 competition.

\end{abstract}
\section{Introduction} \label{intro}

Humanoid robots come with the promise of versatility, allowing to access naturally
human infrastructures thanks to their anthropomorphic design.
Many promising robot architectures were proposed, transitioning from
early designs based on rigid actuators to compliant ones, regaining dynamic
and ensuring safer interactions \cite{ficht2021bipedal}.
However, developing robust locomotion controllers remains an open problem.

The goal of locomotion is for the robot to reach a target pose.
Prior to achieving such a task, the robot has to ensure the security of surrounding humans, and
to preserve its own balance and integrity.
By nature, humanoid robots are underactuated and have to rely on unilateral contacts
with the environment. Because of that, and the high number of degrees of freedom, the equation of
motions governing the robot dynamics is intractable.
To tackle this, simplified models and conservative assumptions are often made, hindering
considerably the robot performances.
Moreover, all the computations have to be performed on-line in an embedded system, where resources are scarce.

Very recently, some work addressed the locomotion problem as a whole in an end-to-end manner, leveraging
DRL techniques.
Lee \textit{et al.} \cite{lee2020learning} proposed a controller for quadruped robots to
navigate on challenging terrains, and deployed on real robots using only proprioceptive information.
Haarnoja \textit{et al.} \cite{haarnoja2023learning} demonstrated impressive soccer skills on bipedal robots, 
but without perception: robot's state estimation is done here by external motion tracking.
Footstep planning is here not considered explicitly, but as a side result of the whole controllers.
Despite exhibiting impressive results, those end-to-end architectures come with complex reward shaping
and costly trainings. Designed for specific tasks, they also arguably lack of versatility and modularity.

\begin{figure}[!t]
    \centering
    \includegraphics[width=\linewidth]{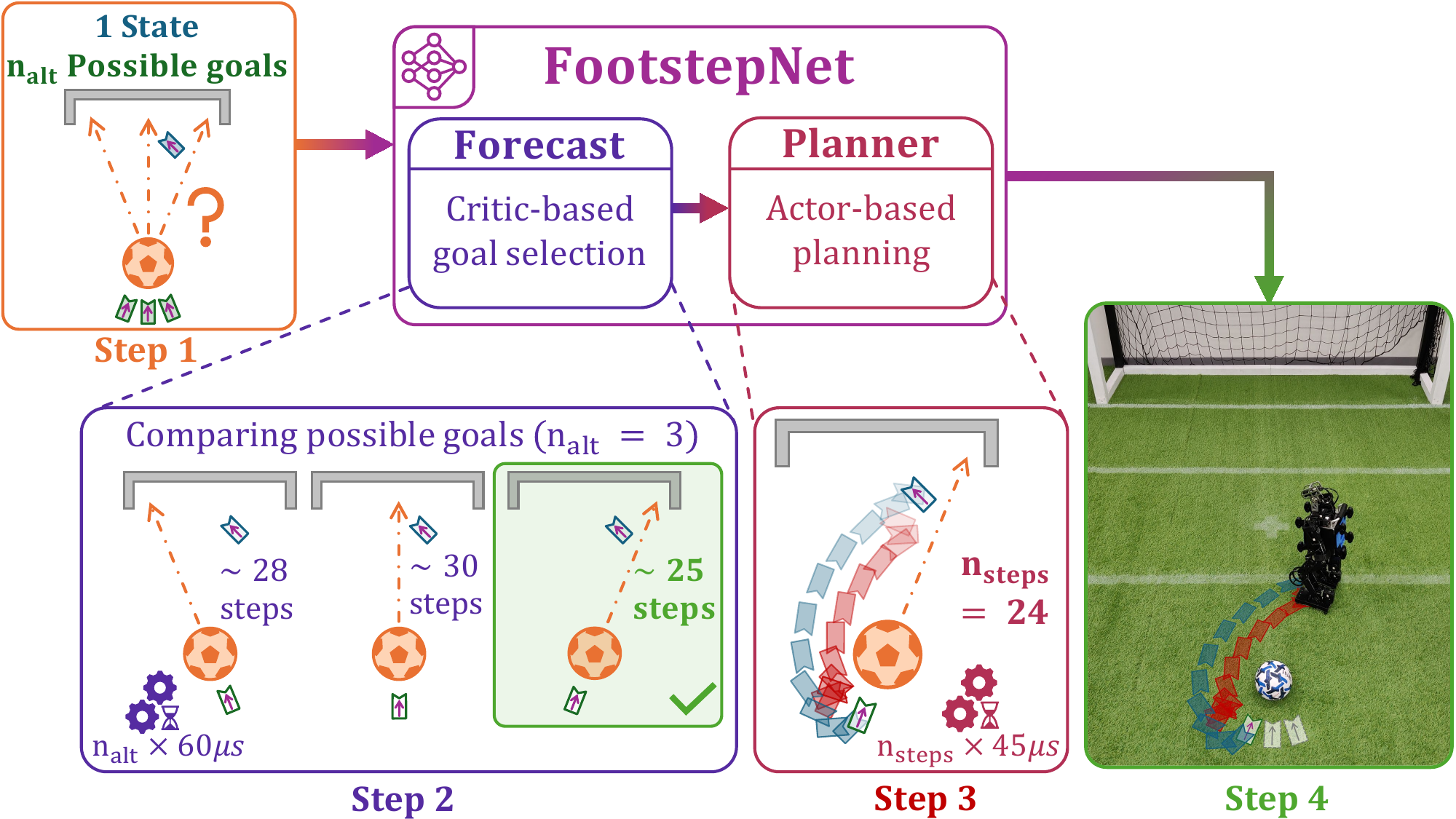}
    \caption{An example of \textit{FootstepNet} use -- \textbf{Step 1}: A bipedal robot must score a goal while minimizing its number of steps. To do this, we arbitrarily choose $n_{alt}$ placement possibilities (here $n_{alt}=3$) which all allow scoring. \textbf{Step 2}: Forecasting allows choosing from the $n_{alt}$ possibilities, the one that requires the fewest steps. \textbf{Step 3}: The planner compute all the steps in order to go to the position chosen by the forecast. \textbf{Step 4}: The step sequence is executed on the real robot.}
    \label{fig:catch-eye}
    \vspace*{-1.6\baselineskip}
\end{figure}

However, in a very classical way, locomotion system design is usually decomposed into sub-problems.
First, the locations (and, optionally, timings) of future contacts are planned in a step focused on footstep planning, which is the central topic of this paper.
Then, using this contact plan, trajectories are computed, typically relying on a simplified model of the robot like
the 3D-LIPM \cite{kajita2003biped}.
Finally, the produced trajectories can then be tracked by a whole-body controller (WBC) \cite{del2015prioritized}.
The boundaries of those sub-problems are arbitrary and were often challenged.

The goal of footstep planning is to find a suitable sequence of footsteps to achieve the desired task.
Some work focuses on finding proved-to-be-\textit{feasible} sequences, which is difficult in complex and
cluttered environments.
In this work, we only consider 2D environment with one obstacle.
We consider the problem of finding the \textit{minimal} sequence.
Even in this simplified form, this optimization problem has no closed-form solution and is challenging
to compute efficiently on-line.
Additionally, navigating in substantially more complex environments can be achieved with a 
global path planner, using intermediate targets for the local planner as presented in \cite{missura2021fast}
and in Fig.~\ref{fig:footstep_planning}.

Footstep planning has been addressed with various approaches for the last 20 years. 
Let us mention at first the bounding box method for 3D environments (see \cite{yoshida2005boundingbox}) when one plans the trajectory of a bounding box for the whole robot, and then the footsteps are deduced from it. It is conservative and does not exploit the advantages of legged locomotion (see also \cite{perrin2012boundingbox}).
Let us also mention Kanoun \textit{et al.} \cite{kanoun2011planning} who proposed to formulate the footstep planning as an inverse kinematics problem, based on \textit{optimization techniques}. The footsteps sequence is represented by an equivalent kinematic chain, where each footstep is made of two prismatic joints and a revolute joint.
Such a problem can be solved iteratively by a gradient descent, which suffers from
inevitable convergence to local optimum.
The most common optimization approaches are based on variants of A* algorithm (see {\em e.g.} \cite{kuffner2001footstep} \cite{chestnutt2005footstep}).

If the possible footsteps are represented as a discrete set, the footstep planning can be
addressed with graph-search algorithms. To that end, many variants of A*\cite{hart1968formal} were
investigated.
Garimort \textit{et al.} \cite{garimort2011humanoid} leveraged the D* Lite \cite{koenig2002d}
algorithm ability to reuse previous searches for replanning.
Hornung \textit{et al.} \cite{hornung2012anytime} proposed to use anytime repairing A* (ARA*) \cite{likhachev2003ara},
where suboptimal initial plan is being refined while navigating, by iteratively reducing
an inflation factor on the heuristic term of the evaluation function.
In those works, the computation time revolves around 1 second, which remains prohibitive
for on-line application. 
A faster search-based approach was proposed by Missura and Bennewitz \cite{missura2021fast}.
In this work, the shortest path is included in the heuristic function in order to abort the search prematurely. 
The authors present a replanning rate of 50Hz, by limiting the computation time to 18ms.
However, other heuristics like the rotate-translate-rotate (RTR) are explicitly added to the formulation.
Even if natural, RTR is still arbitrary and might result in suboptimal behaviour. 
The performance is obtained with assumptions which probably restrict the considered alternatives.
In all those methods,  the design of the footsteps set appears to be an arbitrary and brittle way to
address the trade-off between computation time and suboptimal results.

Finally, \textit{reinforcement learning (RL)} approaches were used.
Hofer and Rouxel \cite{hofer2017operational} proposed a RL-based approach to produce walk orders to approach a soccer
ball. However, they did not consider obstacles.
Meduri \textit{and al.} proposed DeepQ stepper \cite{meduri2021deepq}, a footstep planning based on
DQN algorithm \cite{mnih2013playing}, but their approach was exclusively focused on preserving the
robot stability while tracking a provided velocity task.
Güldenstein \cite{guldenstein2022footstep} proposed a footstep planning method that was compared to an expert method but does not focus on minimizing the number of steps.
In DeepGait \cite{tsounis2020deepgait} by Tsounis \textit{et al.},
quadrupedal locomotion is separated in two DRL subproblems: the Gait Planner (GP) and Gait Controller.
GP addresses a geometrical problem which is the quadruped equivalent of the footstep planning. Even if feasibility
is extensively checked at this stage, it is unclear that the reward function encourages a minimization of the number
of steps because of its complexity.

The first contribution of this paper is \textit{FootstepNet planning}: an efficient \textit{footstep planning} method to navigate in a local environment.
The proposed method can be trained to leverage state-of-the-art DRL techniques and deployed on real robots, 
with very low computational requirements for on-line inference.

The second contribution is \textit{FootstepNet forecasting} : the method's ability to provide an estimated number of footsteps required to reach different candidates of local targets.
Those estimations do not depend on computations of all the required footsteps and can, therefore, produce quick and useful insights
for upstream decision-making (see Fig.~\ref{fig:catch-eye}).

We validate our methodology through simulation outcomes and its successful implementation on 
a small-size humanoid robot during the RoboCup 2023 competition. \cite{allali2024rhoban}.

\section{Problem Statement} \label{PB}

Footstep planning consists in computing the footstep sequence such that the robot can move from an initial position to a target location efficiently and safely, all while avoiding obstacles and adhering to the physical limitations of the robot's mechanics and its environment.
This is a critical aspect of bipedal humanoid robotics.

\begin{figure}[!h]
    \centering
    \includegraphics[width=0.35\textwidth]{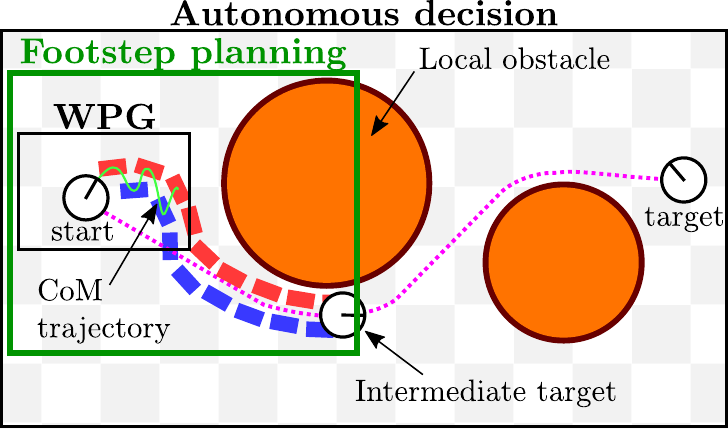}
    \caption{
        Locomotion tasks seen as a hierarchy of problems with different horizons.
        \textit{Autonomous decision} computes a path to navigate globally and an intermediate target to reach.
        \textit{Footstep planning} computes a sequence of footsteps, that ensures the avoidance of the local obstacle.
        \textit{Walk Pattern Generator (WPG)} then computes a Center of Mass (CoM) trajectory
        and use a whole-body controller to follow it.
    }
    \label{fig:footstep_planning}
    \vspace*{-0.65\baselineskip}
\end{figure}

In this paper, we are interested in footstep planning within a two-dimensional (2D) framework as in \cite{Garimort2011} and \cite{Perrin2017}.
By constraining our consideration to the 2D pose of the robot—defined by coordinates ($x$, $y$) and orientation ($\theta$) in a planar domain—we simplify the inherently complex problem of navigation in three-dimensional space. 
This approach allows us to effectively decompose the robot's trajectory into a series of planar movements.

\begin{figure}[!h]
    \centering
    \includegraphics[clip, trim=160px 120px 160px 154px, width=0.5\textwidth]{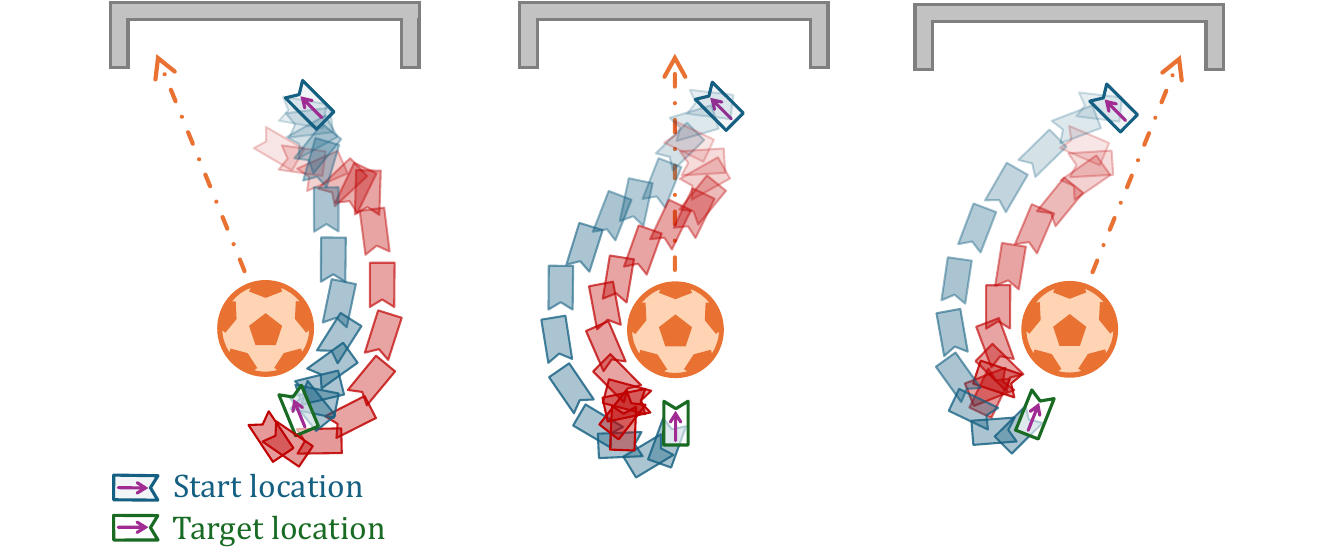}
    \caption{
        Example of footsteps generated by \textit{FootstepNet planning} for the three possible goals of Fig.~\ref{fig:catch-eye} -- The target positions are close to each other, however the generated footsteps to reach them use different complex maneuvers.
    }
    \label{fig:complex_maneuver}
    \vspace*{-1\baselineskip}
\end{figure}

We assume the footstep planning to be part of a broader system, as depicted in Fig.~\ref{fig:footstep_planning}.
In this context, an upstream autonomous decision is made to select the target footstep.
To do so, a global overview of the environment can be used (\textit{e.g.}, using the shortest path with A*-like 
methods).
In our approach, like in \cite{missura2021fast}, we do not consider finding all the footsteps to reach a distant
target.
We rather focus on navigating efficiently in the vicinity of the robot.
Because of the dynamic nature of the environment, as well as slippages and perturbations, replanning become
inevitable and jeopardizes long-term plans. Moreover, long-range navigation is more likely to asymptotically comply
with simple heuristics, such as assuming a constant velocity.
However, with the minimization of the number of footsteps in mind, local navigation can yield complex maneuvers as presented in Fig.~\ref{fig:complex_maneuver}

\section{Background on RL and DRL}

RL considers an agent that interacts with an environment in order to learn the
policy $\pi$ that maximizes the cumulative obtained rewards.
Such a problem can be formulated as a Markov decision process (MDP). An MDP is composed of the tuple
$(\mathcal{S}, \mathcal{A}, P, R)$, where $\mathcal{S}$ is the state space, $\mathcal{A}$ is the action space,
$P$ is the transition function, and $R$ is the reward function.
At every time step $t$, the agent selects an action $a_t \in \mathcal{A}$, follows a transition
from state $s_t$ to state $s_{t+1}$ according to the transition function $P$, and give a reward $r_t = R(s_t,a_t)$.

A deterministic policy $\pi$ maps states to actions such as $\pi : \mathcal{S} \rightarrow \mathcal{A}$. The return $G_t$ is equal to the discounted sum of the rewards. Thus, $G_t = \sum_{i=t}^{T} \gamma^{i-t} R(s_i, a_i)$ where $\gamma \in [0,1]$ is the discount factor, and $T$ is the terminal step. The objective is to find the optimal policy $\pi^*$, which maximizes the average returns. If $\Pi$ is the set of all possible policies, the RL objective is to find $\pi^* = arg\,max_{\pi \in \Pi} \mathbb{E}_\pi [G]$.

To perform this optimization, the RL agent first interacts with the environment and approximates the
action-value, which is the returns expected from a given state-action pair
$Q^\pi(s, a) = \mathbb{E}_\pi [ G_t | s_t = s, a_t = a ]$ (some algorithms approximate
some other closely related quantities). The policy is then updated to maximize the action-value
function. This process is repeated iteratively.

State-of-the art algorithms are based on deep neural networks (DNNs) to approximate both $Q^\pi$ and $\pi$,
yielding the DRL algorithms.
In the case of continuous states and actions, the DRL algorithms are based on the actor-critic
architecture \cite{fujimoto2018addressing} \cite{tuomas2018soft}. In this context, the critic
refers to the action-value $Q^\pi$ which is updated using the temporal difference learning and the
Bellman equation \cite{sutton2018reinforcement} such as:

\begin{equation}
    \label{eq:bellman}
    Q^\pi(s_t,a_t) = r_t + \gamma \mathbb{E}_\pi [Q^\pi(s_{t+1},a_{t+1})].
\end{equation}

The expectation in (\ref{eq:bellman}) is approximated by sampling data obtained from the interaction with the environment.
To that end, the current policy $\pi$ is used with some additional noise to ensure exploration.
The policy can thus be updated by maximizing the policy expected return estimated by the critic:

\begin{equation}
    J(\pi) = \mathbb{E}_\pi [Q^\pi(s_t,a_t)].
\end{equation}

\section{Method} \label{method}






\label{sec:footstep_planning}
\begin{figure}[!t]
    \centering
    \includegraphics[width=0.25\textwidth]{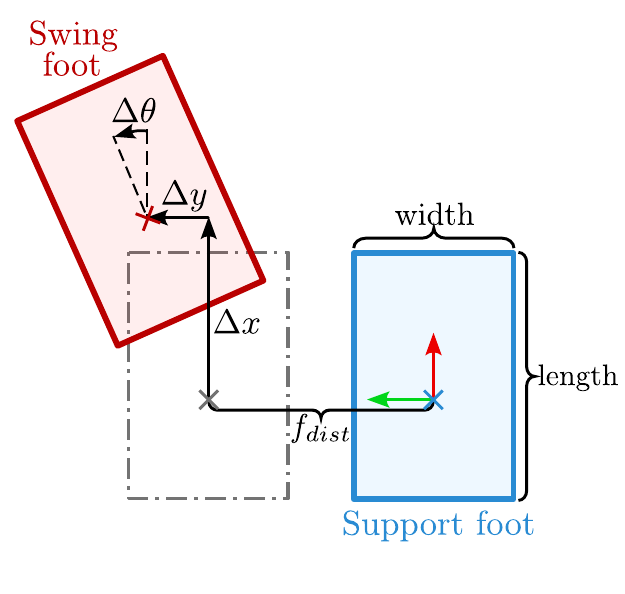}
    \caption{
        Parametrization of a footstep displacement $(\Delta x, \Delta y, \Delta \theta)$.
        The displacement is a pose expressed in the frame of the support foot, with an implicit
        offset of $f_{dist}$ in the $y$ direction.
    }
    \label{fig:displacement_parametrization}
    \vspace*{-1.4\baselineskip}
\end{figure}

We define a footstep as $\phi = (f, x, y, \theta)$, where $f \in \{ \text{left}, \text{right} \}$
indicates a specific foot and $x, y$ and $\theta$ are the position and the orientation of the
foot\footnote{
Unless specified otherwise, all the quantities are expressed in an inertial world frame attached to the ground
}. The robot state can be described with the footstep of its current support
foot $\phi_r = (f_r, x_r, y_r, \theta_r)$. In case of double support, the choice of the support foot is arbitrary.

A footstep displacement $\Delta \phi = (\Delta x, \Delta y, \Delta \theta)$, is parametrized as on Fig.~\ref{fig:displacement_parametrization}.
It describes the pose of the swing foot in the frame of the support foot.
When a support swap occurs, the swing foot becomes the new support foot, producing a new footstep.
A sequence of footsteps can then be built from successive displacements, which defines the trajectory.
\begin{figure*}[!t]
  \includegraphics[clip, trim=0px 141px 0px 145px, width=1\linewidth]{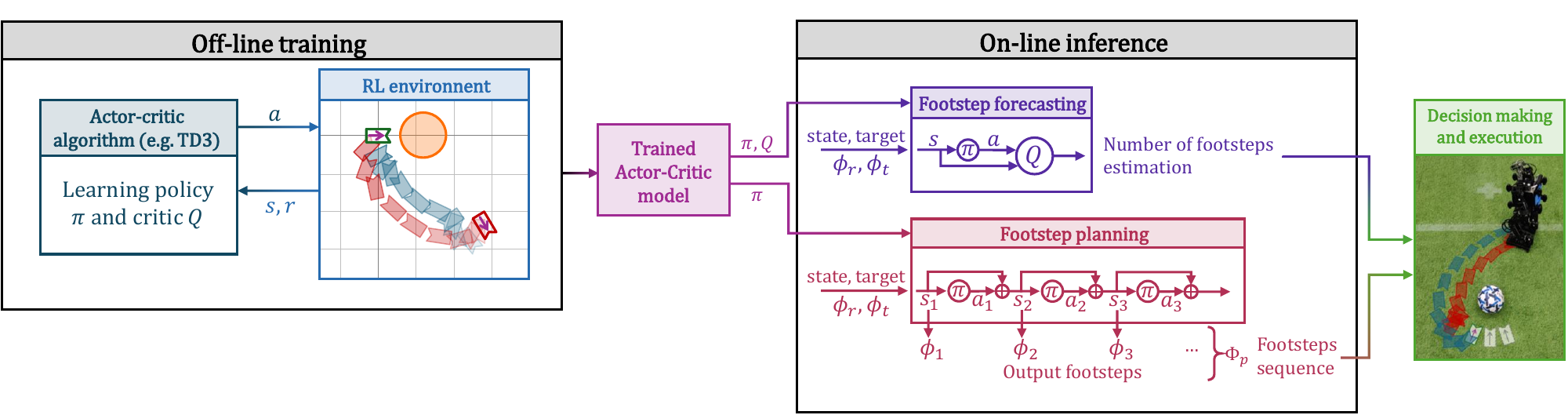}
  \caption{
    \label{fig:framework_overview}
    Overview of the proposed method -- First, offline training is carried out during which the agent learns the policy by interacting with the RL environment. During online inference, we then use the trained networks to, on the one hand, estimate the number of steps using the critic and, on the other hand, to determine the sequence of steps to be performed using the actor.
  }
  \vspace*{-1.2\baselineskip}
\end{figure*}

The displacements are bound in a feasible set $\Delta \phi \in \mathcal{F}$ because the robot has a limited workspace.
Ideally, $\mathcal{F}$ should be able to encompass the ability of the robot to perform the displacement
given its whole-body constraints. In practice, it is approximated with a conservative feasible set.
In this work, $\mathcal{F}$ is a known parameter.
We only assume it to be symmetrical with respect to the sagittal plane of the robot.
However, this assumption is mostly made for state reduction, and can easily be removed with slight adjustments.

An obstacle is defined as $o = (x_o, y_o, \rho)$, where $x_o$ and $y_o$ are the position of the
center of the obstacle and $\rho$ is its radius. A collision between a footstep and an obstacle occurs
if the rectangular support footstep intersects the circular obstacle.

Given a target $\phi_t = (f_t, x_t, y_t, \theta_t)$,
the goal of the \textit{footstep planning} problem is to find a sequence
$\Phi_p = (\phi_r, \phi_2, \dots, \phi_t)$ such that displacements are feasible, and
with minimal length $| \Phi_p |$.
This problem is non-linear because of the possible rotations of the robot.
It also has non-convex constraints because of the obstacle avoidance, but also
possibly because of the shape of $\mathcal{F}$.
For those reasons, there are no known closed-form solutions.

We formulate it as an MDP which has a concise state and action spaces.
This MDP is designed to have a reasonable training time using state-of-the art DRL algorithms.
This allows for a new policy to be computed from the geometrical parameters of the target robot.
Heuristics like RTR \cite{missura2021fast} are no longer needed, this behaviour emerges
from the formulation and the feasible displacements $\mathcal{F}$.
On the other hand, the trained agent has very fast on-board inference time, taking advantage of all modern
hardware acceleration for neural network inferences (Sec. \ref{sec:inference}).

Moreover, taking advantage of the actor-critic architecture, the critic network is also an outcome
of the RL optimization process. Since our reward lead to meaningful return unit (approximating $|\Phi_p|$),
the critic can also be deployed on the robot to perform \textit{footstep forecasting}.
We believe this approach produces a useful building block for the whole locomotion controller.

The MDP formulation is as follows:

\subsubsection{State-space} 

A state $s \in \mathcal{S} = \mathbb{R}^8$ is a tuple:
\begin{equation}
s = (
    \mathbf{1}_{f_r = f_t}, x_t, \sigma(y_t), cos(\theta_t),  \sigma(sin(\theta_t)), x_{o}, \sigma(y_{o}), \rho
),
\end{equation}
where $\mathbf{1}$ is the indicator function : $\mathbf{1}_{f_r = f_t}$ taking the value of $1$
if the robot support foot $f_r$ is the target support foot $f_t$, and $0$ else.

The quantities $x_t, y_t, \theta_t, x_{o}$ and $y_{o}$ are expressed in the support foot
reference frame when included in $s$.
This allows for the current footstep $\phi_r$ to be omitted, reducing the state space dimensionality. 
Moreover, the $\sigma(y)$ operator, defined by $\sigma(y) = y$ if $f_r = right$ and $\sigma(y) = -y$ else,
allows to handle the symmetry of the problem.

\subsubsection{Action-space} 

An action $a \in \mathcal{A} = \mathbb{R}^3$ is a tuple: $a = \Delta \phi$, as specified in Fig.~\ref{fig:displacement_parametrization}.
The actions are clipped to lie in the feasible set $\mathcal{F}$.
After applying back the symmetry operator $\sigma$ on $\Delta y$ and $\Delta \theta$,
such a displacement can be integrated to obtain a new footstep.

\subsubsection{Reward and termination}

The reward function is expressed as:
\begin{equation}
R(s) = -1 - w_1 \delta_p - w_2 \delta_\theta - w_3 \mathbf{1}_{s \in C},
\end{equation}
where $\delta_p$ and $\delta_\theta$ are respectively the distance and absolute orientation error
between the current and the target footstep.
$\mathbf{1}_{s \in C}$ indicates if the current state is in a collision with
the obstacle, $C$ being the set of states in collision.
$0 \le w_1, w_2 \ll 1$ are reward-shaping weights intended to guide the learning
and $w_3$ is a penalty weight.
Every step taken in collision is the equivalent of taking $w_3$ extra steps, which is
prohibitive for $w_3 \gg 1$.
Reaching the target footstep, within a fixed tolerance yields a terminal state
(which is equivalent to a subsequent return of $0$).


\begin{figure*}[!t]
  \centering
  \includegraphics[clip, trim=0px 35px 0px 25px, width=0.83\linewidth]{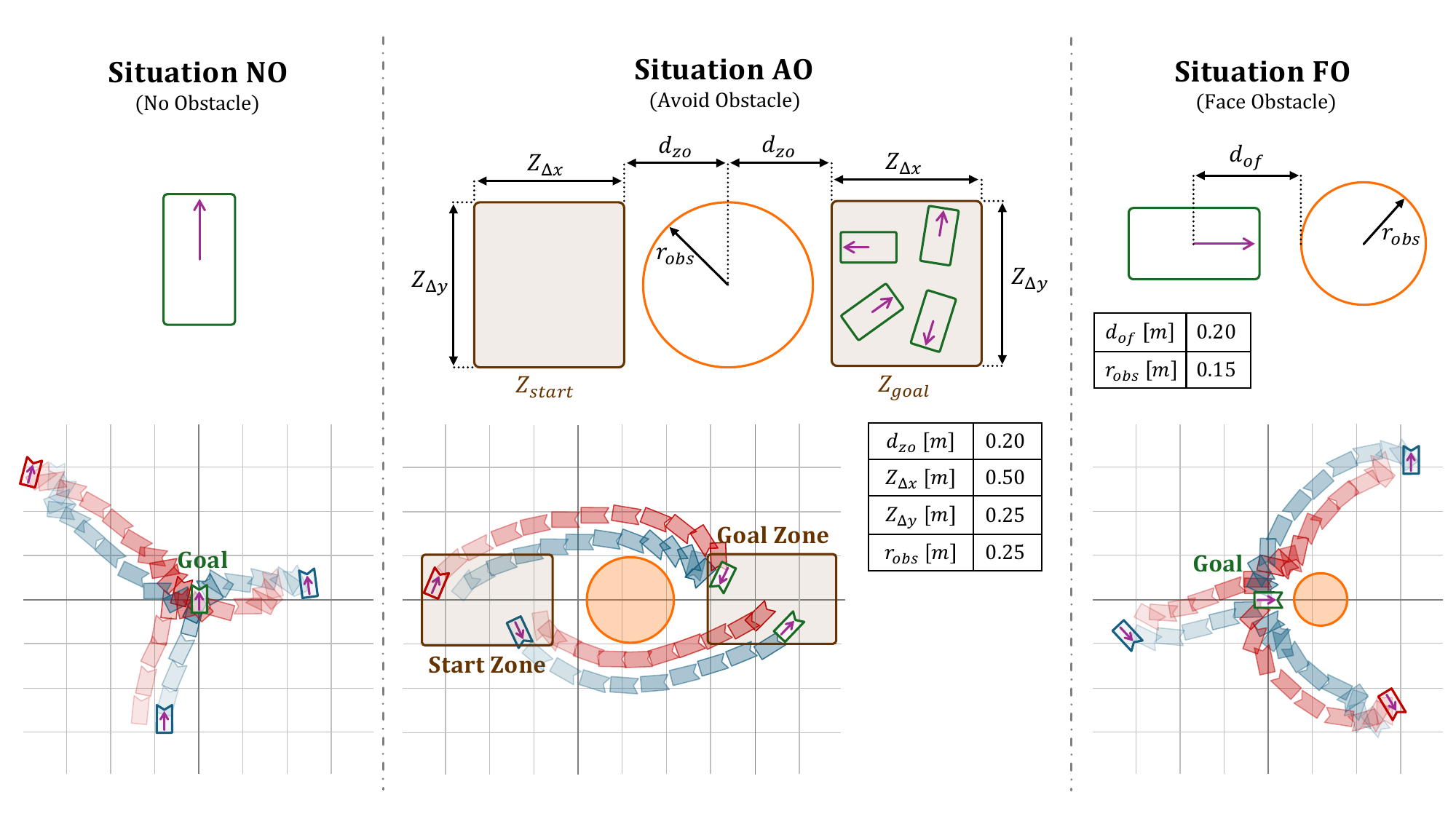}
  \caption{Situations related to Experience 1 : footstep planning -- These situations represent the different scenarios used to compare the performances of FootstepNet planner and ARA* planner. The bottom of the figure represents examples for each of the situations.}
  \label{fig:Exp1}
\end{figure*}

\begin{table*}[!t]
  \centering
  \caption{Results comparison between FootstepNet planner and ARA* planner - 1 000 experiments per column - Footstep sets A and B are respectively the ones based on ASIMO and NAO robots adapted to the Sigmaban platform (SD stands for standard deviation)}
  \begin{tabular}{
      lcccccc
  }
  \toprule
  & \multicolumn{2}{c}{\textbf{Situation NO}} & \multicolumn{2}{c}{\textbf{Situation AO}} & \multicolumn{2}{c}{\textbf{Situation FO}}\\
  \cmidrule(lr){2-3} \cmidrule(lr){4-5} \cmidrule(lr){6-7}
  Obstacle Size [m] & \multicolumn{2}{c}{0} & \multicolumn{2}{c}{0.15} & \multicolumn{2}{c}{0.25} \\
  \midrule
  Discrete footsteps set & {A} & {B} & {A} & {B} & {A} & {B} \\
  Mean(SD) nb of steps - ARA* Planner [steps] & 28.1(7.7) & 27.9(8.1) & 33.8(4.6) & 33.4(5.1) & 27.8(8.2) & 27.9(8.5)\\
  Mean(SD) nb of steps - FootstepNet [steps]& \multicolumn{2}{c}{23.7(6.8)} & \multicolumn{2}{c}{29.8(4.1)} & \multicolumn{2}{c}{24.1(7.8)}\\
  Cases FootstepNet is equal or better [\%] & 100.0 & 99.6  & 97.6 & 97.2 & 99.0 & 99.0\\
  FootstepNet less steps [\%] & 15.91 & 14.93 & 12.30 & 11.0  & 14.97 & 14.79\\
  \bottomrule
  \end{tabular}
  \label{tab:exp1}
  \vspace*{-0.3\baselineskip}
\end{table*}

The return obtained from a given state
can be interpreted as the (negative) approximation of the number of footsteps required to reach the
target.
Given that the critic is an approximation of this return, it can then provide an estimation of
the sequence length $|\Phi_p|$, which is useful for upstream decision-making.
For this reason, the simplicity of the reward function is a key feature of \textit{FootstepNet}.
This approximation is valid if the shaping weights $w_1, w_2$ are small, and if the
discounting factor $\gamma$ is close to $1$.

The sequence of planned footsteps $\Phi_p = (\phi_r, \phi_1, \phi_2, \dots, \phi_H)$
can then be obtained by evaluating recursively the policy with a target horizon $H$.
We call this process a \textit{roll-out} of the policy.
In practice, the size of the horizon $H$ can be selected to produce the relevant number of footsteps
for downstream whole-body planning and control. 
Alternatively, it is possible to apply this \textit{roll-out} fully until the target is reached.
$| \Phi_p |$ then becomes the number of required steps to reach the target.
However, this requires one inference per footstep and is thus costlier than using the critic-based
estimation.

\section{Experiments} \label{exp}

Using the proposed method described in Sec. \ref{method}, an agent dedicated to the footsteps planning was trained.
The computed policy network is used to generate footsteps to reach a given target --\textit{FootstepNet planner}-- and the critic network to forecast the number of steps to the same aim --\textit{FootstepNet forecast}-- (cf. Fig.~\ref{fig:framework_overview})
The main objectives of our experiments were thus to evaluate networks' planning and forecasting performances and to demonstrate the feasibility of the whole pipeline during a real-world scenario : the RoboCup Competition.


\begin{figure*}[!t]
    \centering
    \includegraphics[clip, trim=0px 35px 0px 25px, width=0.83\linewidth]{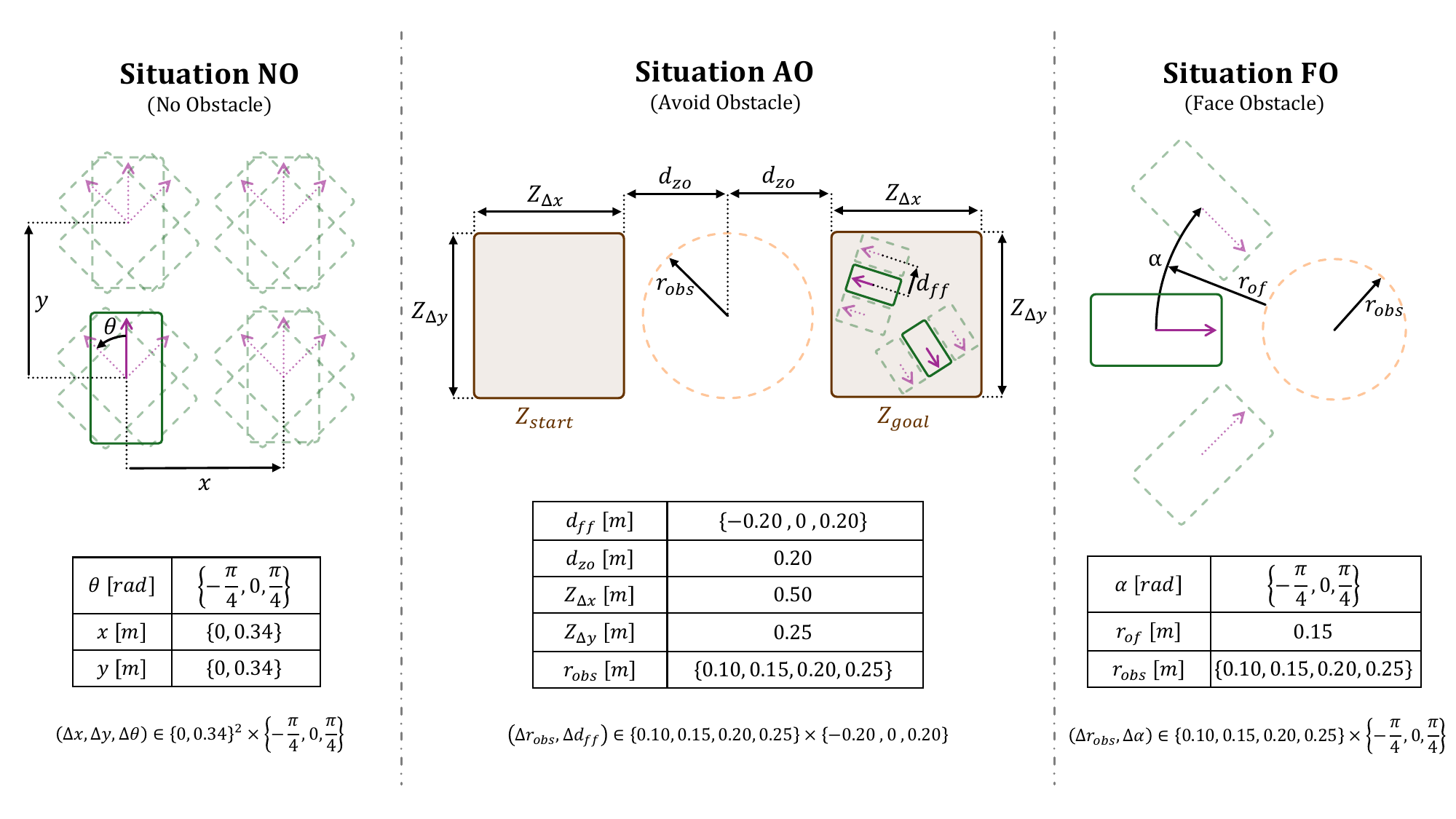}
    \caption{Situations related to Experience 2 : footstep forecasting -- These situations represent the different scenarios used to compare the performances of FootstepNet forecast against FootstepNet planner. For each randomly selected starting pose, the number of steps given by the critic is compared to the roll-out on a set of choices between multiple close/near equivalent targets. The sets of choices for each starting pose are represented at the bottom of the figure.}
    \label{fig:Exp2}
\end{figure*}

\begin{table*}[!t]
    
    \centering
    \caption{Results comparison between FootstepNet forecast and planner - 100 000 experiments per column - the baseline is the number of steps generated by the \textit{FootstepNet planner} to reach the target (SD stands for standard deviation)}
    \resizebox{\textwidth}{!}{%
    \begin{tabular}{
      lccccccccc
    }
    \toprule
     & \multicolumn{1}{c}{\textbf{Situation NO}} & \multicolumn{4}{c}{\textbf{Situation AO}} & \multicolumn{4}{c}{\textbf{Situation FO}}\\
    \cmidrule(lr){2-2} \cmidrule(lr){3-6} \cmidrule(lr){7-10}
    Obstacle Size [m] & \multicolumn{1}{c}{0} & 0.10 & 0.15 & 0.20 & 0.25 & 0.10 & 0.15 & 0.20 & 0.25 \\
    \midrule
    Mean estimation relative error [\%] & 6.02  & 4.13 & 4.55 & 5.10 & 5.95 & 4.97 & 5.11 & 5.67 & 5.93\\
    Best case mean(SD) nb. of steps - Actor [steps] & 20.7(7.4)  & 26.4(4.2) & 26.8(4.1) & 27.3(4.1) & 27.9(4.2) & 21.4(7.4) & 21.5(7.3) & 21.4(7.4) & 21.6(7.4) \\
    Worst case mean(SD) nb. of steps - Actor [steps] & 30.4(7.7)  & 29.6(4.4) & 30.0(4.2) & 30.7(4.2) & 31.6(4.5) & 30.3(7.6) & 30.5(7.5) & 30.9(7.8) & 31.3(7.9) \\
    Critic-based erroneous-choice ratio [\%] & 4.64  & 5.24 & 6.68 & 9.65 & 13.26 & 0.37 & 0.59 & 0.74 & 0.76 \\
    Extra steps taken for erroneous choice [\%] & 0.58 & 0.46 & 0.64 & 0.88 & 1.31 & 0.05 & 0.07 & 0.10 & 0.11\\
    Steps improvement choosing best choice vs worst [\%] & 46.36  & 12.00 & 12.20 & 12.48 & 13.13 & 41.67 & 42.29 & 43.90 & 44.87 \\
    \bottomrule
    \end{tabular}%
    }
    \label{tab:exp2}
    \vspace*{-0.5\baselineskip}
\end{table*}

\subsection{Setup}

\subsubsection{Parameters} \label{sec:parameters}

Our experiments were conducted on the Sigmaban platform \cite{allali2024rhoban}, a kid-size humanoid robot (0.7m, 7.7kg). It is therefore its characteristics and capabilities that 
were used as parameters of the RL environment in order to train FootstepNet. 
Each foot is 0.14m long and 0.08m wide, and the distance between the two feet is 0.15m.

In this work, we assume the displacements of each foot to be bound in an ellipsoid ensuring that
\begin{equation}
\label{eq:displacement_constraint}
\lVert
\begin{bmatrix} \frac{\Delta x}{\Delta x_{max}} & \frac{\Delta y}{\Delta y_{max}} & \frac{\Delta \theta}{\Delta \theta_{max}}
\end{bmatrix}^T
\rVert_2 \leq 1,
\end{equation}
where $\Delta x_{max}$, $\Delta y_{max}$ and $\Delta \theta_{max}$ are the maximum allowed displacements in the $x$, $y$ and $\theta$ 
directions and $\lVert \cdot \rVert$ denotes the Euclidean norm.
Because of the forward/backward asymmetry of the robot, the maximum displacements in the $x$ direction is 
different for forward (0.08m) and backward (0.03m) displacements. The maximum displacement in the $y$ ($\pm$0.04m) 
and $\theta$ ($\pm$20°) axis remains the same for both directions.
Similar approach was used by \cite{missura2021fast} in order to reduce extreme combination of multiple directions.
Indeed, the ellipsoid shape embraces the robot's workspace in a less conservative way than its bounding
box counterpart.

The local area of the RL environment in which the robot can evolve is a 4x4m square.
The tolerances for reaching the target goal, triggering episode termination, were set to 0.05m and 5°.
Episodes were truncated after 90 steps to ensure periodic reset of the environment and augment state-space
exploration at the beginning of the training.

All these parameters can be changed to correspond to another bipedal robot, a wider workspace or a different set of constraints.


\subsection{Footstep planning} \label{sec:footstep_planning_exp}

\subsubsection{Training} \label{sec:training}

The agent was trained using TD3 \cite{fujimoto2018addressing}, one of the state-of-the art DRL algorithm as implemented in \cite{raffin2019stable} using a PC with an Intel® Core™ i7-9700K CPU, an NVIDIA GeForce RTX 2070 GPU, 32Go of RAM and an SSD.

Both neural networks (for the actor and the critic) are two-layer perceptrons,
featuring hidden layers of 400 and 300 neurons respectively, as detailed in \cite{fujimoto2018addressing}. The model employs an initial learning rate of $10^{-3}$, which is linearly annealed during training. For exploration, normal noise with a 10\% standard deviation is introduced, and is also linearly decayed. Additionally, the training process adopts a LeakyReLU activation function, a batch size of 256 and a discount factor of 0.98.
The output layer employs a Tanh activation function to guarantee that the output values adhere to the confines of the action space.
The model is trained for 10 million steps, which corresponds to 4 hours. This training time can
likely be significantly reduced by finer hyperparameters tuning.
Using a sparse reward with Hindsight Experience Replay (HER) \cite{andrychowicz2017hindsight} was also considered but without better success than our dense reward.




\subsubsection{On-board inference} \label{sec:inference}

From the actor-critic trained model, the actor was extracted to create the \textit{FootstepNet planner} and the combination 
of the actor and one of the Q-Networks of TD3 was used to create the \textit{FootstepNet forecast}. In order to try to exploit 
the maximum capacities of our computational power during runtime, we used the open source OpenVINO™ Runtime \cite{openvino_2023}.
We only used the computer's CPU.
The on-board computer of the Sigmaban platform used for inference is an Intel NUC running with Ubuntu 22.04 and composed of 
an Intel® Core™ i5-7260U CPU, 8Go of RAM and an SSD.
The mean inference time to generate one foot pose with \textit{FootstepNet planner} is $45\mu s$, $60\mu s$ are required by 
\textit{FootstepNet forecast} to predict the total number of steps to reach a target from a given state.



The first experiment is dedicated to compare the performances of \textit{FootstepNet planner} against the state-of-the-art footstep 
planner ARA* \cite{likhachev2003ara} \cite{hornung2012anytime}. We used implementation made by J. Garimort and al. \cite{Garimort2011} 
in the ROS footstep planner package\footnote{\url{https://wiki.ros.org/footstep_planner}}.
In order to do so, three different scenarios were created (cf. Fig.~\ref{fig:Exp1}) : 
\\

\begin{itemize}
    \item \textbf{Situation NO} : No obstacle, the robot has to reach a fixed target without any obstacle. The starting poses are randomly chosen in the 4x4m local area defined in Sec.\ref{sec:parameters}
    \item \textbf{Situation AO} : Avoid obstacle, two zones are created to force the robot to avoid a fixed size obstacle (radius of 0.15m). The starting and goal poses are randomly assigned within these zones.
    \item \textbf{Situation FO} : Face obstacle, the robot has to face an obstacle to reach a fixed target. The starting poses are defined as for the first situation. \\
    Indeed, facing a table, an object to manipulate or a human to interact with are common tasks in bipedal robotics.
\end{itemize}

The main difference between both planners is that ARA* is based on anytime heuristic search, necessitating a discrete footsteps set, while FootstepNet leverages a DRL algorithm for a continuous footsteps set. The neeed to select a discrete footsteps set, a complex task, is obviated in our approach. 
The discrete footstep sets used for this experiment are the ones defined in \cite{Garimort2011} and \cite{hornung2012anytime}, tested on ASIMO and NAO robots. They were adapted to the range of Sigmaban for the purpose of the experiment.

The results presented in Table \ref{tab:exp1} demonstrate that the \textit{FootstepNet planner} consistently surpasses the performance of ARA* in all tested scenarios. Indeed, in the worst case, the RL agent is equal or better in 97.2\% of the experiments. Moreover, we fixed the maximum search time for ARA* to 10s for each target to reach, which is a reasonable time given that, according to \cite{hornung2012anytime}, it takes 5s to find a near-optimal path. Compared to that, the execution time of \textit{FootstepNet planner} is $45\mu s$ per footstep, which is negligible compared to ARA*.



\subsection{Footstep forecasting} \label{sec:forecasting_exp}


The second experiment aims to validate the accuracy of \textit{FootstepNet forecast} compared to the \textit{roll-out} of the policy.
The forecasting predicts the number of steps to reach the target from a given state. The \textit{roll-out} is the number of steps generated by the \textit{FootstepNet planning} until the target is reached.
The same scenarios as in the first experiment were used (cf. Fig.~\ref{fig:Exp2}) with the addition of four different obstacle sizes for the \textbf{AO} and \textbf{FO} situations.
However, for each randomly selected starting pose, the critic was compared to the roll-out on a set of choices between multiple close/near equivalent targets.
Table \ref{tab:exp2} shows that the forecasting is nearly as effective as the \textit{roll-out} to select the best target to reach among these near equivalent ones (eg. For facing a 0.15m obstacle, the critic-based erroneous ratio is only 0.59\%).
According to the low mean estimation relative error (4.55\% for avoiding a 0.15m obstacle), we can deduce that \textit{FootstepNet forecast} is also able to accurately predict the number of steps to reach a target.


\subsection{Deployment on a kid-size humanoid during RoboCup 2023}
\label{sec:exp_robocup}

RoboCup is a large international robotics competition, happening almost every year since 1997. One of its league is humanoid soccer, where teams of robots face each other trying to score goals. 

\begin{figure}[!h]
    \centering
    \includegraphics[width=0.85\linewidth]{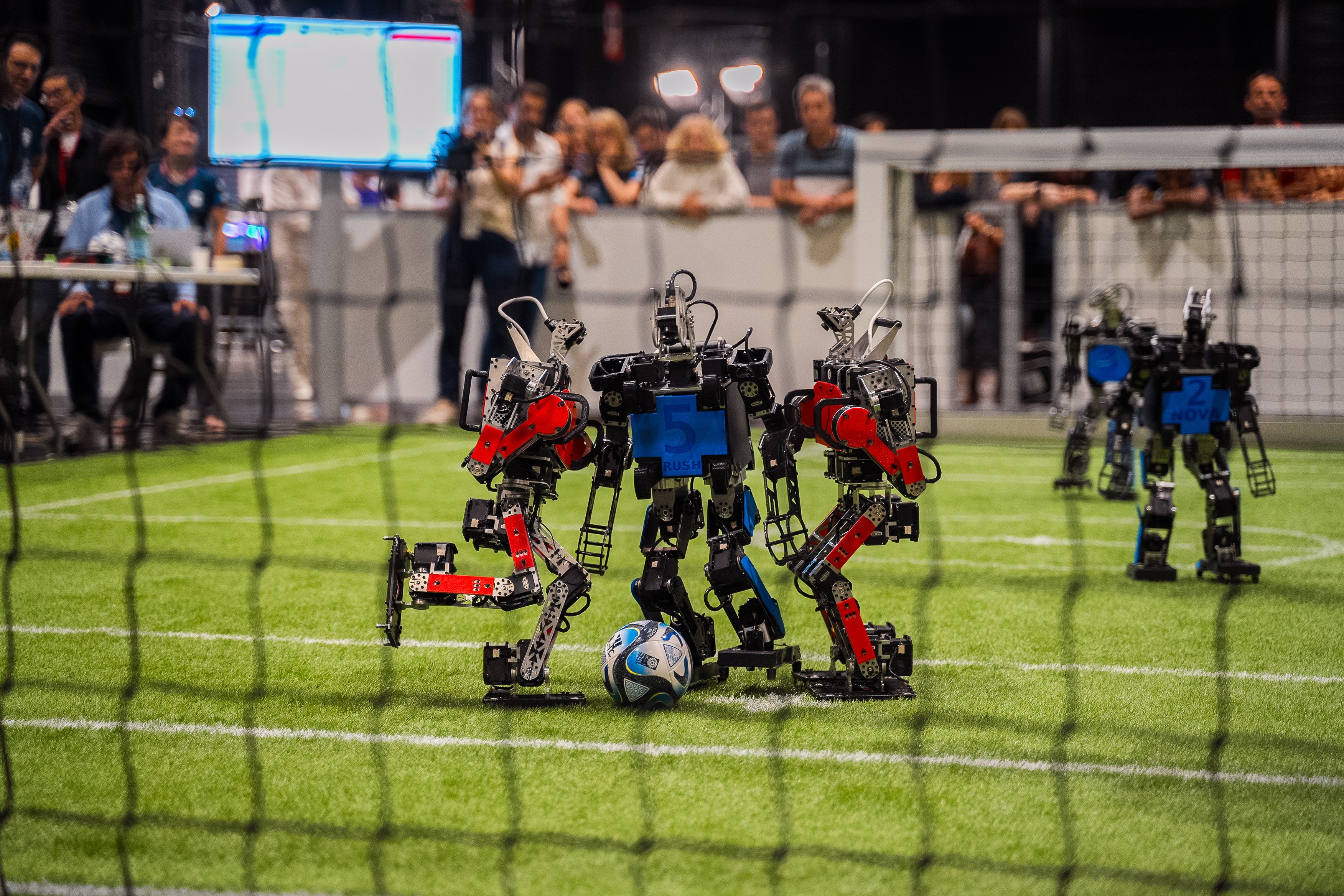}
    \caption{Sigmaban robots (in blue) during a RoboCup 2023 soccer match}
    \label{fig:robocup}
    \vspace*{-0.9\baselineskip}
\end{figure}

\begin{figure}[!h]
    \centering
    \includegraphics[width=0.75\linewidth]{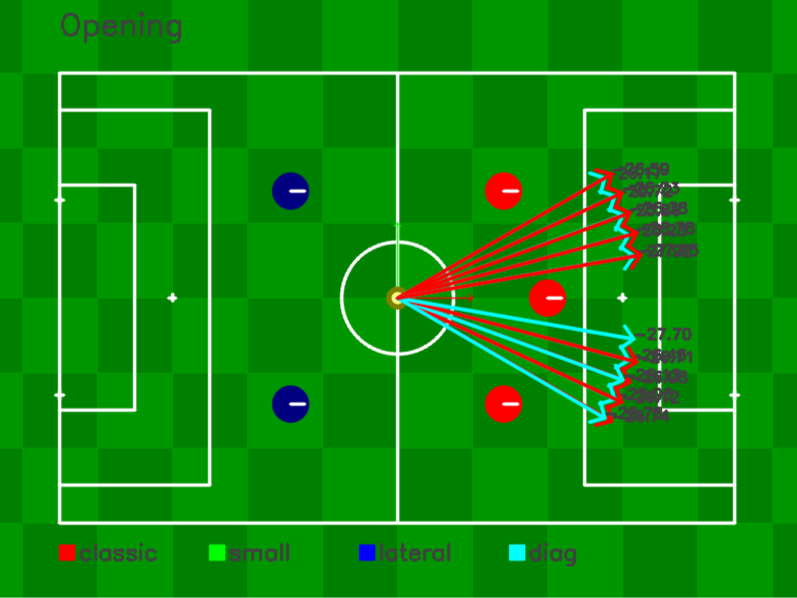}
    \caption{An example of a situation in our strategy viewer where the robot has to choose a position to kick the ball (in yellow). Allies are in blue and opponents in red. Arrows represent the possible positions of the ball after the kick.}
    \label{fig:strategy}
\end{figure}

The robots are custom-made and fully autonomous, carrying their battery and computational power. For the 2023 edition, we deployed FootstepNet in Sigmaban, competing in the kid-size category. In a competition setup, it is crucial to take as few footsteps as possible to be as fast as possible. In particular, many maneuvers are necessary around the ball to get in position to kick (see Fig.~\ref{fig:catch-eye}).
Let us mention that this task, i.e. reach a position as quickly as possible by walking, is natural beyond the context of soccer, and would be useful in many other applications. Let us also mention that the competition context requires a high level of reliability.
The footsteps were re-planned periodically with an horizon of $H=5$ footsteps, yielding a constant computation time
of $225\mu s$.
Footsteps are then passed to the downstream whole-body planner to plan the CoM trajectory with a scheme similar to
\cite{dimitrov2008implementation}.
To decide the target kick and placement, an upstream strategy/decision-making module was designed. Relevant information (position of the ball, allies and opponents) was used to select target poses, 
using an estimated time-to-goal score. Since several poses were often equivalent, FootstepNet forecasting was then used to make the final choice (Fig.~\ref{fig:strategy}).

We scored 95 goals, took 2 and won the competition. FootstepNet planned all the footsteps taken and helped extensively in fine decision-making thanks to forecasting. 
This was achieved on-board with low computational power, releasing precious CPU resources for other tasks.

Additionally, we provide a video\footnote{\url{https://youtu.be/sLu-LHdO6Mk}} about FootstepNet and demonstrates its application on a Sigmaban robot.

\section{Conclusion} \label{CCL}





In conclusion, the comprehensive evaluation and deployment of FootstepNet have underscored its effectiveness and efficiency as a planner in bipedal robotics, particularly in comparison to the state-of-the-art ARA* planner. Through experimentation under various scenarios, including obstacle navigation and target reaching, FootstepNet has consistently demonstrated superior performance, achieving equal or better results in the vast majority of tests while boasting significantly lower execution times. The utilization of DRL with a continuous set of footsteps not only streamlines the planning process but also obviates the need for selecting a discrete footsteps set, a notable advantage over traditional methods. Additionally, the accurate forecasting capability of FootstepNet, as evidenced in both experimental setups and real-world competition scenarios such as RoboCup 2023, highlights its potential for enhancing decision-making in robotics, enabling quick and reliable movements essential for success in dynamic environments.

We explained in Sec. \ref{method} that we approximate the action space of the feet by an ellipsoid clipping. However, this approximation could be enhanced by considering the true action space of the feet, which remains a significant challenge to accurately determine. Additionally, our method could be extended to accommodate more complex local environments, including those with non-circular obstacles.

Overall, despite this, FootstepNet represents a significant step forward in the domain of footstep planning, combining speed, efficiency, and accuracy in a manner not previously achieved by existing planners, to our knowledge. Its success in both controlled experiments and competitive environments attests to its utility and potential for broader applications.

The open-source code for this project is available on our GitHub repository\footnote{\url{https://github.com/Rhoban/footstepnet_envs}} for further use and contributions.

\bibliographystyle{ieeetr}
\bibliography{biblio}

\end{document}